# Filling Knowledge Gaps in a Broad-Coverage Machine Translation System[*]

Kevin Knight, Ishwar Chander, Matthew Haines, Vasileios Hatzivassiloglou,
Eduard Hovy, Masayo Iida, Steve K. Luk, Richard Whitney, Kenji Yamada
USC/Information Sciences Institute
4676 Admiralty Way
Marina del Rey, CA 90292

## Abstract

Knowledge-based machine translation (KBMT) techniques yield high quality in domains with detailed semantic models, limited vocabulary, and controlled input grammar. Scaling up along these dimensions means acquiring large knowledge resources. It also means behaving reasonably when definitive knowledge is not yet available. This paper describes how we can fill various KBMT knowledge gaps, often using robust statistical techniques. We describe quantitative and qualitative results from JAPANGLOSS, a broad-coverage Japanese-English MT system.

## 1 Introduction

Knowledge-based machine translation (KBMT) techniques have led to high quality MT systems in circumscribed problem domains [Nirenburg et al., 1992] with limited vocabulary and controlled input grammar [Nyberg and Mitamura, 1992]. This high quality is delivered by algorithms and resources that permit some access to the meaning of texts. But can KBMT be scaled up to unrestricted newspaper articles? We believe it can, provided we address two additional questions:

1. In constructing a KBMT system, how can we acquire knowledge resources (lexical, grammatical, conceptual) on a large scale?

2. In applying a KBMT system, what do we do when definitive knowledge is missing?

There are many approaches to these questions. Our working hypotheses are that (1) a great deal of useful knowledge can be extracted from online dictionaries and text; and (2) statistical methods, properly integrated, can effectively fill knowledge gaps until better knowledge bases or linguistic theories arrive.

When definitive knowledge is missing in a KBMT system, we call this a *knowledge gap*. A knowledge gap may be an unknown word, a missing grammar rule, a missing piece of world knowledge, etc. A system can be designed to respond to knowledge gaps in any number of ways. It may signal an error. It may make a default or random decision. It may appeal to a human operator. It may give up and turn over processing to a simpler, more robust MT program. Our strategy is to use programs (sometimes statistical ones) that can operate on more readily available data and effectively address particular classes of KBMT knowledge gaps. This gives us robust throughput and better quality than that of default decisions. It also gives us a platform on which to build and test knowledge bases that will produce further improvements in quality.

Our research is driven in large part by the practical problems we encountered while constructing JAPANGLOSS, a Japanese-English newspaper MT system built at USC/ISI. JAPANGLOSS is a year-old effort within the PANGLOSS MT project [Nirenburg and Frederking, 1994; NMSU/CRL et al., 1995], and it participated in the most recent ARPA evaluation of MT quality [White and O'Connell, 1994].

## 2 System Overview

This section gives a brief tour of the JAPANGLOSS system. Later sections address questions of knowledge gaps on a module-by-module basis.

Our design philosophy has been to chart a middle course between "know-nothing" statistical MT [Brown et al., 1993] and what might be called "know-it-all" knowledge-based MT [Nirenburg et al., 1992]. Our goal has been to produce an MT system that "knows something" and has a place for each piece of new knowledge, whether it be lexical, grammatical, or conceptual. The system is always running, and as more knowledge is added, performance improves.

Figure 1 shows the modules and knowledge resources of our current translator. With the exceptions of JUMAN and PENMAN, all were constructed new for JAPANGLOSS. The system includes components typical of many KBMT systems: a syntactic parser driven by a feature-based augmented context-free grammar, a semantic analyzer that turns parse trees into candidate interlinguas, a semantic ranker that prunes away meaningless interpretations, and a flexible generator for ren-dering interlingua expressions into English.

---

[*]This work was supported in part by the Advanced Research Projects Agency (Order 8073, Contract MDA904-91-C-5224) and by the Department of Defense. Vasileios Hatzivassiloglou's address is: Department of Computer Science, Columbia University, New York, NY 10027. All authors can be reached at *lastname*@isi.edu.

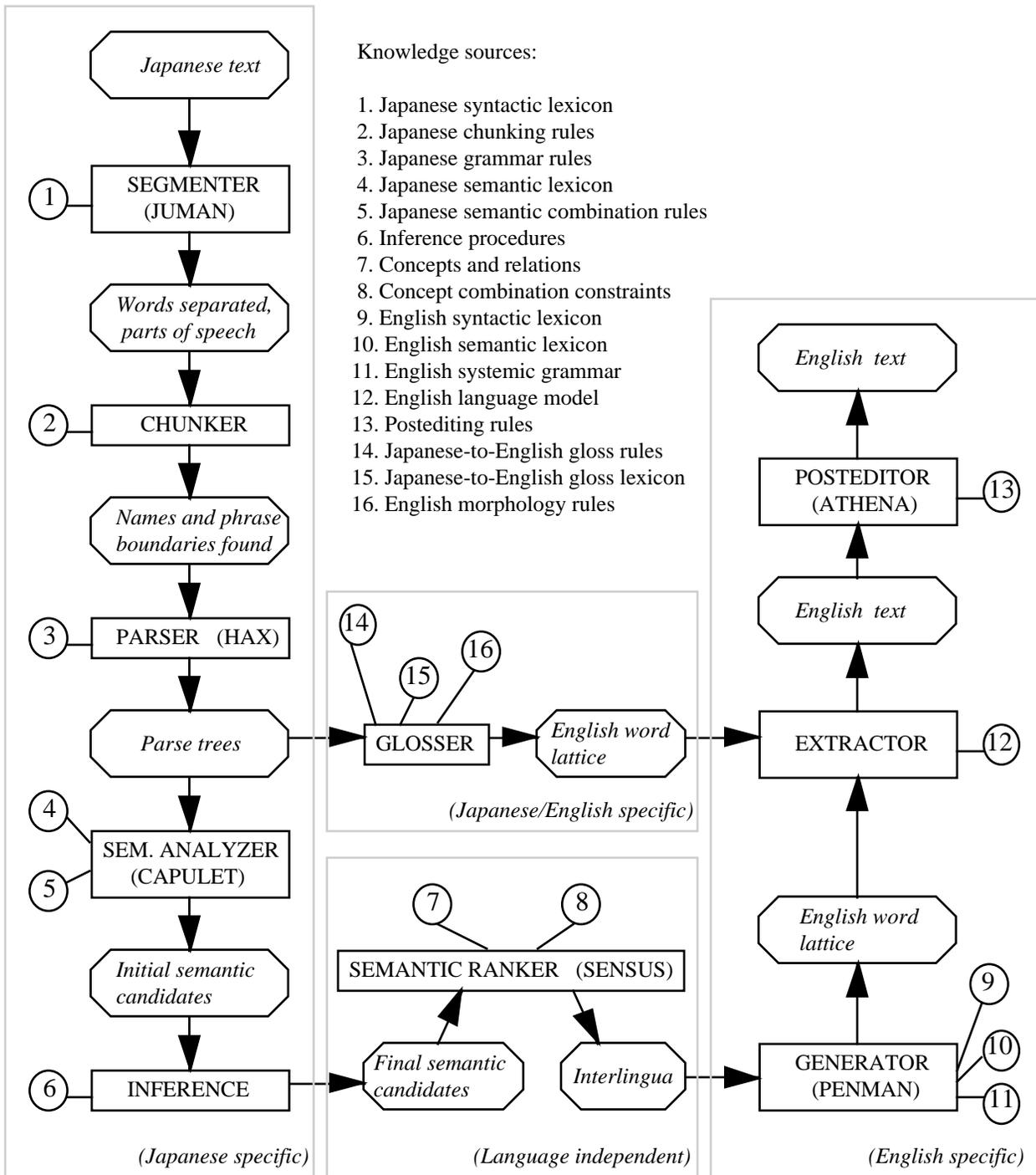

Figure 1: The JAPANGLOSS Machine Translation System: Modules and Knowledge Sources

Because written Japanese has no inter-word spacing, we added a word segmenter. We also added a pre-parsing (chunking) module that performs several tasks: (1) dictionary-based word chunking, (2) recognition of personal, corporate, and place names, (3) number/date recognition, and (4) phrase chunking. This last type of chunking inserts new strings into the input sentence, such as `BEGIN-NP` and `END-NP`. The syntactic grammar is written to take advantage of these barriers, prohibiting the combination of partial constituents across phrase boundaries. VP and S chunking allow us to process the very long sentences characteristic of newspaper text, and to reduce the number of syntactic analyses. We also employ a limited inference module between semantic analysis and ranking. One of the tasks of this module is to take topic-marked entities and insert them into particular semantic roles.

The processing modules are themselves independent of particular natural languages—they are driven by language-specific knowledge bases. The interlingua expressions are also language independent. In theory, it should be possible to translate interlingua to and from any natural language. In practice, our interlingua is sometimes close to the languages we are translating. Here is an example interlingua expression produced by our system:

```
(h-709 / HAVE-AS-A-GOAL
   :SENSER    (c-710 / COMPANY-BUSINESS
                :Q-MOD (n-711 / NEW-VIRGIN))
   :PHENOMENON (f-712 / FOUND-LAUNCH
                :TEMPORAL-LOCATING
                   (c-713 / CALENDAR-MONTH
                      :MONTH-INDEX 2)
                :AGENT c-710)
   :THEME c-710)
```

The source for this expression was a Japanese sentence whose literal translation is something like: *as for new company, there is plan to establish in February.*

The tokens of the expression are drawn from our conceptual knowledge base (or ontology), called SENSUS. SENSUS is a knowledge base skeleton of about 70,000 concepts and relations; see [Knight and Luk, 1994] for a description of how we created it from online resources. Note that the interlingua expression above includes a syntactic relation `Q-MOD` ("quality-modification"). While our goal is to replace such relations with real semantic ones, we realize that semantic analysis of adjectives and nominal compounds is an extremely difficult problem, and so we are willing to pay a performance penalty in the near term.

In addition to SENSUS, we have large syntactic lexical resources for Japanese and English (roughly 100,000 stems for each), a large grammar of English inherited from the PENMAN generation system [Penman, 1989], a large semantic lexicon associating English words with any given SENSUS concept, and hand-built chunking and syntax rules for the analysis of Japanese. Sample rules can be found in [Knight *et al.*, 1994].

We have not achieved high throughput in our semantic analyzer, primarily because we have not yet completed a large-scale Japanese semantic lexicon, i.e., mapped tens of thousands of common Japanese words onto conceptual expressions. For progress on our manual and automatic attacks on this problem, see [Okumura and Hovy, 1994] and [Knight and Luk, 1994]. So we have encountered our first knowledge gap—what to do when a sentence produces no semantic interpretations? Other gaps include incomplete dictionaries, incomplete grammar, incomplete target lexical specifications, and incomplete interlingual representations.

## 3   Incomplete Semantic Throughput

While semantic resources are under construction, we want to be able to test our other modules in an end-to-end setting, so we have built a shorter path from Japanese to English called the *glossing* path.

We made very minor changes to the semantic interpreter and rechristened it the glosser. Like the interpreter, the glosser also performs a bottom-up tree walk of the syntactic parse, assigning feature structures to each node. But instead of semantic feature structures, it assigns glosses, which are potential English translations encoded as feature structures. A sample gloss looks like this:

```
((GLOSS
  ((OP1
    ((OP1 (*OR* "a" "an" "the" "*empty*"))
     (OP2 "national police agency")
     (OP3 (*OR* "+plural" "*empty*"))))
   (OP2
    ((OP1
      ((OP1 (*OR* "a" "an" "the" "*empty*"))
       (OP2 (*OR* "plan" "objective"))
       (OP3 (*OR* "+plural" "*empty*"))))
     (OP2 "of")
     (OP3
      ((OP1 (*OR* "a" "an" "the" "*empty*"))
       (OP2 ((OP1 ((OP1
                    ((OP1 "gun")
                     (OP2 (*OR* "engraving tool"
                                "knife"
                                "saber" "sword"))))
                   (OP2 (*OR* "mood" "divisor"
                              "doctrine" "process"
                              "way" "method"
                              "rule" "law"))))
              (OP2 (*OR* "alteration" "amendment"))))
       (OP3 (*OR* "+plural" "*empty*"))))))))))
```

In this structure, the `OP1`, `OP2`, etc. features represent sequentially ordered portions of the gloss. All syntactic and glossing ambiguities are maintained in disjunctive (`*OR*`) feature expressions.

We built three new knowledge resources for the glosser: a bilingual dictionary (to assign glosses to leaves), a gloss grammar (to assign glosses to higher constituents, keying off syntactic structure), and an English morphological dictionary.

The output of the glosser is a large English word lattice, stored as a state transition network. The typical network produced from a twenty-word input sentence has 100 states, 300 transitions, and billions of paths. Any path through the network is a potential English translation, but some are clearly more reasonable than others. Again we have a knowledge gap, because we do not know which path to choose.

Fortunately, we can draw on the experience of speech recognition and statistical NLP to fill this gap. We built a language model for the English language by estimating bigram and trigram probabilities from a large collection of 46 million words of Wall Street Journal material.[1] We smoothed these estimates according to class membership for proper names and numbers, and according to an extended version of the *enhanced Good-Turing method* [Church and Gale, 1991] for the remaining words. The resulting tables of conditional probabilities guide a statistical extractor, which applies a version of the N-best beam search algorithm [Chow and Schwartz, 1989] to identify an ordered set of "best" paths in the word lattice (i.e., the set of English sentences that are most likely according to our model). Due to heavy computational and memory requirements, we have not yet completed the trigram version of our model. But even when only bigrams are used, comparing the statistical extractor to random path extraction reveals the power of statistics to make reasonable decisions, e.g.:

```
(random extractor)

"...planned economy ages is threadbare..."

(bigram extractor)

"...planned economy times are old..."
```

The statistical model of English gives us much better glosses. A full description of our glossing system, including our use of feature unification, appears in [Hatzivassiloglou and Knight, 1995]. Interestingly, we originally built the statistical model to address knowledge gaps in generating English from interlingua, not glossing. We return to these gaps (and the statistical model) in Section 8.

## 4 Incomplete Dictionaries

Lexical incompleteness causes problems for both semantics and glossing. Numbers and dates are typical unknown words; we take a finite-state approach to recognizing and translating these. Another important class of unknown words comprises katakana loanwords and foreign proper names, which are represented in Japanese with approximate phonetic transliterations. For example, *kurinton* should be translated as *Clinton*, and *suteppaa mootaa* as *stepper motor*. The knowledge-based approach is to pack our dictionaries with every possible name and technical term, but this approach quickly leads to diminished returns. To plug the rest of the gap, we have again applied statistical techniques, this time to

---

[1] Available from the ACL Data Collection Initiative, CD ROM 1.

English spelling. Given a Japanese transliteration like *kurinton,* we seek an English-looking word likely to have been the source of the transliteration. "English-looking" is defined in terms of common four-letter sequences. To produce candidates, we use another statistical model for katakana translations, computed from an automatically aligned database of 3,000 katakana-English pairs. This model tells us that Japanese *ku* is sometimes a transliteration of English *c* (especially at the beginning of a word), sometimes of *ck* (especially at the end), sometimes of *cu*, etc., with associated probabilities. Another extractor program delivers a reasonable transliteration, for example, preferring *Clinton* over *Kuleentn*.

## 5 Incomplete Grammatical Analysis

Unlike spoken language, newspaper text is generally grammatical. However, it is frequently ungrammatical with respect to our knowledge resources at any given time. Newspaper sentences are very long, and every new text contains some unusual syntactic construction that we have not yet captured in our formal grammar. We also encounter problems with non-standard punctuation as well as wrong segmentation, part-of-speech tagging, and phrase chunking. For these reasons, we made an early decision to do all parsing, semantic interpretation, and glossing in a bottom-up fashion, dealing independently with sentence fragments when need be. But because Japanese and English word orders are so different, a fragmentary parse usually leads to a bad translation.

One method of overcoming these difficulties is to ensure a full parse with statistical context-free parsing. Low probability rules like ADV → ADJ guarantee that a full parse will be returned [Yamron *et al.*, 1994]. We avoided this technique because we wanted to keep our feature-based grammar, in order to have fine-tuned control over the structures we accept and assign. Inspired by [Lavie, 1994], we turned to word skipping as a grammatical gap-plugger. This approach seeks out the largest grammatical subset of words in a sentence, with the hope that the skipped words are peripheral to the core meaning of the sentence (or perhaps simply stray punctuation marks). Rather than port the LR-parsing-based techniques of [Lavie, 1994] to our bottom-up chart parser, we developed heuristic techniques for spotting suspicious words and dropping them. Our most general method is to automatically process large sets of parsed and unparsed sentences, looking for statistical differences between the two sets. For example, if a part-of-speech bigram appears frequently in unparsed sentences, this is a clue that one of the two words (or both) should be dropped. We also use grammar-specific heuristics such as: don't drop a single noun from a noun sequence, drop phrasal boundaries only in pairs (and only when the internal material is completely parsed), etc. Skipping raises our full-parse rate from 40% to 90%.

## 6 Incomplete Semantic Constraints

This section and the next return to incompleteness in semantics. KBMT systems usually model world knowledge as a collection of "hard" constraints that any in-terlingua must satisfy. Unsatisfactory semantic candidates are pruned away, leaving a single sensible interpretation. In many cases, however, semantic constraints are too strong, and all interpretations are ruled out. Preference semantics [Wilks, 1975] was devised to handle just this problem. Within the KBMT tradition, role restrictions can be augmented with `relaxable-to` statements [Carlson and Nirenburg, 1990], as in: the `agent` of a `say-event` must be a `person`, or `relaxable-to` an `organization`.

Our approach has been to further soften the impact of semantic constraints. We assign a score (between 0 and 1) to any interlingua fragment, as follows. We first extract all stated relations between entities. To each relation, we assign scores based on domain constraints and range constraints in the conceptual knowledge base. There are five possible heuristic scores, depending on the suitability of the role filler: satisfies basic constraint (1.0); satisfies relaxed constraint, but is not mutually disjoint from concepts satisfying basic constraint (0.8); satisfies relaxed constraint but *is* mutually disjoint from basic constraint (0.25); satisfies neither basic nor relaxed constraint (0.05); is mutually disjoint from concepts satisfying basic or relaxed constraint (0.01). Scores for all of the relations are multiplied to yield an overall score. No interlingua expression is ever assigned a zero score.

This approach is like language modeling, but here the basic unit is the relational/conceptual n-gram (e.g., `<eat, patient, worm>`) rather than the word n-gram. And our scoring is based on hand-built inherited constraints rather than data. Still, we can view the conceptual knowledge base *as a device that assigns a priori probabilities to interlingua fragments*, in analogy to how our language model assigns probabilities to English strings.

This analogy makes it possible to describe KBMT in a statistical framework. Direct statistical MT systems [Brown *et al.*, 1993] use a noisy channel model in which a human is assumed to be speaking English, but the signal is corrupted, and out comes Japanese. Bayes' Theorem is used to retrieve the original, uncorrupted signal:

$$\hat{E} = \underset{E}{\operatorname{argmax}} \ P(E) \cdot P(J \mid E)$$

That is, the best English translation is the sentence E that maximizes the *a priori* probability of E times the probability that if E were the original signal, it would have been corrupted into the Japanese sentence J. Estimating the probability distributions P(E) and P(J | E) allows us to rank translations.

A noisy channel model of KBMT adopts a different model of a human, one in which he hears English, but by the time it gets into his head, it has been corrupted into Interlingua. When he speaks, the Interlingua signal is further corrupted into Japanese. We can model this process statistically as:

$$\hat{I} = \underset{I}{\operatorname{argmax}} \ P(I) \cdot P(J \mid I)$$

$$\hat{E} = \underset{E}{\operatorname{argmax}} \ P(E) \cdot P(\hat{I} \mid E)$$

Now there are four probability distributions to estimate, one of which is P(I), exactly the *a priori* probability of an Interlingua expression described above.

While the full statistical model of KBMT is not used in JAPANGLOSS, it is useful to view what we do (and don't do) in this light. For example, P(I | E) suggests a model of English generation that explicitly shies away from ambiguous E's, because they spread probabilities thinly across several I's. For example, if $\hat{I}$ is `RENT-TO(SPEAKER,HOUSE,?)`, it is better to say *I'm renting my house to someone* rather than the correct but ambiguous *I'm renting my house*. Most language generation systems focus on accurate renditions rather than unambiguous ones, although as Pereira and Grosz [1993, p. 12] remark, this is changing:

> "The coextensiveness of natural language perceivers and producers both enables and requires the language generator to reason about the generated language in reflective terms: 'how would I react if I heard what I am about to say?' This reflective aspect of language generation is essential at all levels of generation ..."

## 7 Incomplete Interlingua Expressions

Even with full semantic throughput and accurate ranking of interlingua candidates, a full account of text meaning is beyond the state of the art. It is easy to record verb tenses, but difficult to make the inferences required to lay out stated (and unstated) events on a time line. If the source language is Japanese, we have additional problems: no articles (*a, an, the*), no overt singular or plural marks, no agreement constraints, omitted subjects, no marked future tense, and so on. In KBMT, we need some semantic representation of these things in order to generate languages like English. This leads researchers to envision microtheories [NMSU/CRL *et al.*, 1995] of time, space, reference, and so on. Many such microtheories are not yet available, however, so we have another knowledge gap to fill.

We can focus on the problem of definite and indefinite reference, which manifests itself as article selection (*a, an, the*) in Japanese-English MT. Rather than handle this problem during semantics, we postpone the solution—not only until English generation, but until *after* generation, in an automatic postediting step. Our posteditor inserts articles into article-free English text [Knight and Chander, 1994]. It was trained on 80 megabytes of English and performs with an accuracy of 81%. The training was done with decision trees rather than n-grams, so that we could flexibly integrate long distance features that typically control the selection of articles. Human posteditors can achieve 96%, so we still have more features to explore. Viewing article selection as a postediting step (independent of the Japanese source text) means that we can attack the problem statistically without the need for a large parallel corpus of Japanese and English. Our automatic posteditor has applications outside of MT, such as improving English text written by native Japanese speakers (or Chinese, Russian, etc.).

## 8 Incompleteness in Generation

We have encountered several classes of knowledge gaps in large-scale generation. Large-scale in our case means roughly 70,000 concepts/relations and 91,000 roots/phrases. First, we must anticipate incompleteness in the input specification, as described in the last section. Second, lexical syntactic specification may be incomplete—does verb V take a nominal direct object or an infinitival complement? Third, collocations may be missing. Knowledge of collocations has been successfully used in the past to increase the fluency of generated text [Smadja and McKeown, 1991]. In particular, such knowledge can be crucial for selecting prepositions (*on the field* versus *in the end zone*) and other forms. And fourth, dictionaries may not mark rare words and generally may not distinguish between near-synonyms. We constantly strive to augment our knowledge bases and lexicons to improve generation, but we also want to plug the gap with something reasonable.

Our approach is one of bottom-up *many-paths* generation [Knight and Hatzivassiloglou, 1995], in analogy to bottom-up all-paths parsing. If the generator cannot make an informed decision (lexical selection, complement type, spatial preposition, etc.), it pursues all possibilities. This is in contrast to many industrial-strength generators [Tomita and Nyberg, 1988; Penman, 1989] that never backtrack and usually default their difficult choices. Other generators [Elhadad, 1993] do backtrack but still use default or random schemes for making new selections.

Our generator packs all paths into efficient word lattices as it goes, and the final output is also a word lattice. Because the generator produces the same data structure as the glosser, we can select a final output using the same extractor and language model described in Section 3. This approach combines the strengths of knowledge-based generation (e.g., generally grammatical lattice paths, long-distance agreement, parallel conjoined expressions) with statistical modeling (e.g., local dependencies, lexical constraints, common words and collocations). As in Section 3, we can compare random path extraction with n-gram extraction:

(random extractor)

"The new companies will have as a purpose
 launching at February."

(bigram extractor)

"The new company plans to establish
 in February."

(random extractor)

"A subsidiary of the Perkin Elmer Co. on
 the Japan bears majority of the stock."

(bigram extractor)

"A subsidiary of Perkin Elmer Co. in Japan
 bears a majority of the stock."

As the generator becomes more knowledgeable, its output lattices become leaner, and dependence on automatic statistical selection is reduced. However, the statistical component is still useful as long as uncertainty remains, and improvements in language modeling will continue to have a big effect on overall performance.

## 9 Conclusion

We have reported progress on aspects of the JAPANGLOSS newspaper MT system. In particular, we have described the integration of statistical and heuristic methods into a KBMT system. While these methods are not a panacea, they offer a way to fill knowledge gaps until better knowledge bases become available. Furthermore, they offer a way of rationally prioritizing manual tasks: if a statistical method solves your problem 90% of the time, you may not want to invest in a knowledge base. In other cases, statistics may offer only a small benefit over random or default choices; then, a more careful analysis is called for.

## Acknowledgments

We would like to thank Yolanda Gil and the IJCAI reviewers for helpful comments on a draft of this paper.

## A JAPANGLOSS MT Output

### A.1 Interlingua-Based Sample

```
Citizen Watch announced on the eighteenth to
establish a joint venture with Perkin Elmer Co.,
a major microcircuit manufacturing device
manufacturer, and to start the production of the
microcircuit manufacturing device.
  The new company plans a launching in February.
  The subsidiary of Perkin Elmer Co. in Japan
bears a majority of the stock, and the production
of the dry etching device that is used for the
manufacturing of the microcircuit chip and the
stepper is planned.
```

### A.2 Glosser-Based Sample

```
The "gun possession" penalties important--national
police agency plans of gun sword legal reform.
  The national police agency defend policies that
change some of gun sword law that put the brakes on
the proliferation of the 31st gun.  Change plans
control recovery plan of the wrong owning step
currency--make the decision by the cabinet meeting
of the middle of this month in three pieces support
estimate of the filing in this parliament.
```